%% file: iclr2022_conference.tex
\documentclass[letterpaper]{article} 
\usepackage[dvipsnames,table]{xcolor} 
\usepackage{iclr2022_conference,times}
\iclrfinalcopy

\usepackage{hyperref}
\usepackage{url}

\usepackage[utf8]{inputenc} 
\usepackage[T1]{fontenc}    
\usepackage{booktabs}       
\usepackage{amsfonts}       
\usepackage{nicefrac}       
\usepackage{microtype}      
\usepackage{amsmath,amssymb,amsfonts}
\usepackage{amsthm}
\usepackage[linesnumbered,ruled,vlined]{algorithm2e}
\usepackage{todonotes}
\newtheorem{assumption}{Assumption}
\newtheorem{theorem}{Theorem}
\newtheorem{corollary}{Corollary}
\newtheorem{lemma}[theorem]{Lemma}
\newtheorem{definition}{Definition}
\usepackage{makecell}
\usepackage{multirow}
\usepackage{enumitem}
\usepackage{wrapfig}

\hypersetup{
	colorlinks,
	citecolor=blue,
	linkcolor=red,
	breaklinks=true,
	urlcolor=black
}

\title{A Reduction-Based Framework for Conservative Bandits and Reinforcement Learning}

\author{Yunchang Yang\thanks{equal contribution} \\
Center for Data Science, Peking University\\
\texttt{yangyc@pku.edu.cn} \\
\And
Tianhao Wu\footnotemark[1] \\
University of California, Berkeley \\
\texttt{thw@berkeley.edu} \\
\And 
Han Zhong\footnotemark[1] \\
Center for Data Sience, Peking University \\
\texttt{hanzhong@stu.pku.edu.cn} \\
\And 
Evrard Garcelon, Matteo Pirotta, Alessandro Lazaric\\
Facebook AI Research\\
\texttt{\{evrard, pirotta, lazaric\}@fb.com} \\
\AND
Liwei Wang \\
Key Laboratory of Machine Perception, MOE, \\
School of Artificial Intelligence, Peking University \\
International Center for Machine Learning Research, \\
Peking University \\
\texttt{wanglw@cis.pku.edu.cn}\\
\And 
Simon S. Du \\
University of Washington\\
\texttt{ssdu@cs.washington.edu}
}

\newcommand{\para}[1]{\noindent{\bf #1.}}

\iclrfinalcopy 
\begin{document}

\maketitle
\allowdisplaybreaks
\begin{abstract}
   \input{abs.tex}

\end{abstract}

\section{Introduction}
\label{sec:intro}
\input{intro.tex}

\subsection{Related Work}
\label{sec:rel}
\input{related}

\section{Preliminaries}
\label{sec:pre}
\input{preliminaries}

\section{General Framework For Conservative Exploration}
\label{sec:framework}
\input{framework}

\section{Regret Lower Bound for Conservative Exploration}
\label{sec:lb}
\input{lb.tex}
\section{Upper Bounds}
\label{sec:ub}
\input{ub}

\section{Conclusion}
\label{sec:con}
\input{conclusion.tex}

\section*{Acknowledgements}
Liwei Wang was supported by National Key R\&D Program of China (2018YFB1402600), Exploratory Research Project of Zhejiang Lab (No. 2022RC0AN02), BJNSF (L172037). Project 2020BD006 supported by PKUBaidu Fund.

\bibliography{ref}
\bibliographystyle{iclr2022_conference}

\newpage{}
\appendix
\section{Details about Bandits and RL}
\label{app:pre}
\input{formulation}

\input{appendix}
\end{document}

%% file: abs.tex
We study bandits and reinforcement learning (RL) subject to a conservative constraint where the agent is asked to perform at least as well as a given baseline policy.
This setting is particular relevant in real-world domains including digital marketing, healthcare, production, finance, etc.
In this paper, we present a reduction-based framework for conservative bandits and RL, in which our core technique is to calculate the necessary and sufficient budget obtained from running the baseline policy.
For lower bounds, we improve the existing lower bound for conservative multi-armed bandits and obtain new lower bounds for conservative linear bandits, tabular RL and low-rank MDP, through a black-box reduction that turns a certain lower bound in the nonconservative setting into a new lower bound in the conservative setting. 
For upper bounds, in multi-armed bandits, linear bandits and tabular RL, our new upper bounds tighten or match existing ones with significantly simpler analyses. 
We also obtain a new upper bound for conservative low-rank MDP.

%% file: intro.tex


This paper studies online sequential decision making problems such as bandits and reinforcement learning (RL) subject to a conservative constraint.
Specifically, the agent is given a reliable \emph{baseline policy} that may not be optimal but still satisfactory.
In conservative bandits and RL, the agent is asked to perform nearly as well (or better) as the baseline policy at all time.
%
This setting is a natural formalization of many real-world problems such as digital marketing, healthcare, finance, etc.
For example, a company may want to explore new strategies to maximize profit while simultaneously maintaining profit above a fixed baseline at any time, in order not to be bankrupted.
See~\citep{wu2016conservative} for more discussions on the motivation of the conservative constraint. 


Analogously to the non-conservative case, conservative bandit/RL problems also require us to balance exploration and exploitation carefully. Meanwhile, to ensure the obtained policies outperform the baseline policy, we need to provide a tractable approach to keep the exploration not too aggressive. Solving these two problems simultaneously is the key challenge in conservative bandits and RL.

Existing work proposed algorithms for different settings, including bandits~\citep{wu2016conservative,kazerouni2016conservative,garcelon2020improved,Katariya2019interleaving,zhang2019contextual,du2020onesizefitsall,wang2021conservative}
and tabular RL~\citep{garcelon2020conservative}.
However, lower bound exists only for the multi-armed bandit (MAB) setting ~\citep{wu2016conservative}, and there is no lower bound for other widely-adopted settings, such as linear bandits, tabular Markov Decision Process (MDP) and low-rank MDP. In Section~\ref{sec:rel}, we provide a more detailed discussion of the related work.

For each of the different settings considered in the literature (i.e., multi-armed bandits, linear bandits, tabular MDPs), existing approaches rely on ad-hoc algorithm design and analysis of the trade-off between the setting-specific regret analysis and the conservative constraint. 
Furthermore, it is hard to argue about the optimality of the proposed algorithms because it would require clever constructions of the hard instances to prove the non-trivial regret lower bounds under the conservative constraint.

\subsection{Our Contributions}
In this paper, we address these limitations and make significant progress in studying the general problem of online sequential decision-making with conservative constraint.
We propose a \emph{unified framework} that is generally applicable to online sequential decision-making problems.
The common theme underlying our framework is to calculate the necessary and sufficient budget required to enable non-conservative exploration. Such a budget is obtained by running the baseline policy (cf. Section~\ref{sec:framework}). With the new framework, we obtain a novel upper bound on tabular MDPs, which improves the previous result. And we prove a new upper bound on low-rank MDPs. Also, we derive the first lower bounds for linear bandits, tabular and low-rank MDPs, which shows that our upper bound is tight. 

\noindent \textbf{Lower Bounds.} For any specific problem (e.g., multi-armed bandits, linear bandits), our framework \emph{immediately} turns a minimax lower bound of the non-conservative setting to a non-trivial lower bound for the conservative case (cf. Section~\ref{sec:lb}).
We list some examples to showcase the power of our framework for lower bounds.
Full results are given in Table~\ref{tab:regret.bounds}.
\begin{itemize}[noitemsep,topsep=0pt,parsep=0pt,partopsep=0pt,leftmargin=0.1in]
	\item We derive a novel lower bound for \emph{multi-armed bandits} that works on a wider range of parameters than the one derived in \citep{wu2016conservative}. In particular, our lower bound shows a more refined dependence on the value of the baseline policy.  
	\item We derive the \emph{first} regret lower bound for conservative exploration in \emph{linear bandits}, \emph{tabular MDPs} and \emph{low-rank MDPs}. These results allow to establish or disprove the optimality of the algorithms currently available in the literature.
\end{itemize}

We emphasize our technique for deriving lower bounds is simple and generic, so we believe it can be used to obtain lower bounds for other problems as well.

\textbf{Upper Bounds.}
Our novel view of conservative exploration can also be used to derive high probability regret upper-bounds.
When the suboptimality gap $\Delta_0$ and the expected return $\mu_0$ of the baseline policy are known, we show that the \textsf{Budget-Exploration} algorithm (Alg.~\ref{alg:budgetfirst}) attains minimax optimal regret in a wide variety of sequential decision-making problems, when associated to any minimax optimal non-conservative algorithm specific to the problem at hand. 
In the more realistic (and challenging) scenario where $\Delta_0$ and $\mu_0$ are unknown, we show how to simply convert an entire class of algorithms
with a sublinear non-conservative regret bound into a conservative algorithms with a sublinear regret bound. 
We obtain the following results, full details are given in Table~\ref{tab:regret.bounds}.
\begin{itemize}[noitemsep,topsep=0pt,parsep=0pt,partopsep=0pt,leftmargin=0.1in]
	\item In the MAB setting, we obtain a regret upper-bound that matches our refined lower-bound, thus improving on existing analysis. In the linear bandit setting, we match existing bounds that are already minimax optimal.
	\item In the RL setting, we provide two novel results. First, we provide the first minimax optimal result for tabular MDPs, improving over~\citep{garcelon2020conservative}. Second, we derive the first upper bound for conservative exploration in low-rank MDPs. Our bound matches the rate of existing non-conservative algorithms though it is not minimax optimal. How to achieve minimax optimality in low rank MDPs is an open question even in non-conservative exploration.
\end{itemize}
Again, our reduction technique is simple and generic, and can be used to obtain new results in
previously unstudied settings, like we did for low-rank MDPs.

\begin{table}[t!]
	\centering
	\footnotesize
		\renewcommand{\arraystretch}{1.5}
		\begin{tabular}[0.6\textwidth]{|c|c|c|}
		\hline
		Setting & Lower Bound & \makecell{Upper Bound}\\
		\hline\hline
		\multirow{2}{*}{\makecell{Multi-armed\\ bandits}} &

			\cellcolor[HTML]{F5F5F5}
			$\Omega\left(\sqrt{AT}+ \frac{A\Delta_0}{\alpha\mu_0\left(\alpha\mu_0+\Delta_0\right)}\right)$

		&
			\cellcolor[HTML]{F5F5F5}
				$\widetilde{O}\left(\sqrt{AT} + \frac{A\Delta_{0}}{\alpha \mu_{0}(\alpha\mu_{0} + \Delta_{0})}\right)$
		\\
		\cline{2-3}
		&\makecell{$\Omega(\sqrt{AT} + \frac{A}{\alpha \mu_{0}})$\\{\scriptsize \citep{wu2016conservative}
		\footnotemark[2]}}
		&\makecell{$\widetilde{O}\left(\sqrt{AT} + \frac{A}{\alpha \mu_{0}} \right)$\\{\scriptsize \citep{wu2016conservative}}}
		\\
		\hline
			\makecell{Linear\\ bandits}
		&\cellcolor[HTML]{F5F5F5}
		$\Omega\left(d\sqrt{T}+ \frac{d^{2}\Delta_0}{\alpha \mu_0 \left(\alpha \mu_0 + \Delta_{0}\right)}  \right)$
		&\cellcolor[HTML]{F5F5F5}
		\makecell{$\widetilde{O}\left(d\sqrt{T} + \frac{d^{2}\Delta_{0}}{\alpha \mu_{0}(\alpha\mu_{0} + \Delta_{0})}\right)$\\{\scriptsize This work and \citep{kazerouni2016conservative, garcelon2020improved}}}
		\\
		\cline{2-3}
		\hline

		\multirow{2}{*}{\makecell{Tabular\\ MDPs}}
		&\cellcolor[HTML]{F5F5F5}
		&\cellcolor[HTML]{F5F5F5} $\widetilde{O}\left(\sqrt{H^3SAT} +  \frac{SAH^{3}\Delta_0}{\alpha \mu_0 \left(\alpha\mu_0 + \Delta_0\right)}\right)$\\
		\cline{3-3}
		&
		\multirow{-2}{*}{
		 	\cellcolor[HTML]{F5F5F5}
		 	$\Omega\left(\sqrt{H^3SAT}+  \frac{SAH^{3}\Delta_0}{\alpha \mu_0 \left(\alpha_0 + \Delta_0\right)}\right)$
		}
		&\makecell{$\widetilde{O}\left(\sqrt{H^3SAT} + \frac{S^{2}AH^{5}\Delta_{0}}{\alpha \mu_{0}(\alpha \mu_{0} + \Delta_{0})}\right)$\\{\scriptsize \citep{garcelon2020conservative}}}
		\\
		\hline
		\makecell{Low Rank\\ MDPs}
		&\cellcolor[HTML]{F5F5F5}
		$\Omega\left(\sqrt{d^{2}H^3T}+ \frac{d^2H^3\Delta_0}{\alpha \mu_0 \left(\alpha\mu_0 + \Delta_0\right)}\right)$
		&\cellcolor[HTML]{F5F5F5}
		$\widetilde{O}\left(\sqrt{d^{3}H^4T} + \frac{d^3H^4\Delta_0}{\alpha \mu_0 \left(\alpha\mu_0 + \Delta_0\right)}\right)$
		\\
		\hline
		\end{tabular}
	\vspace{0.04in}
	\caption[]{
		Comparison of bounds for conservative decision-making. \textbf{Our contributions are reported in grey cells.}
		We denote by $T$  the number of rounds the agent plays (episodes in RL), $\alpha$ the conservative level,
		$\mu_0$ the expected return of the baseline policy\footnotemark[3], 
		$\Delta_0$ the suboptimality gap of the baseline policy, $A$ the number of actions (or arms),
		$S$ the number of states and $d$ the feature dimension.
	    The upper bounds hold both in the case $\Delta_0$ and $\mu_0$ are unknown since the lack of knowledge changes the regret only by a constant multiplicative factor (cf. Section~\ref{sec:ub}).
	}
\label{tab:regret.bounds}

\end{table}

\subsection{Main Difficulties and Technique Overview} 

\subsubsection{Lower Bounds}
The only lower bound for conservative exploration is by \cite{wu2016conservative} who followed a classical approach in the bandit literature. 
They constructed a class of hard environments and used an information-theoretic argument to prove the lower bound. 
Construction of hard environments is highly non-trivial because one needs to incorporate the hardness from the conservative constraint.
It is also non-trivial to generalize \cite{wu2016conservative}'s lower bound to other settings such as conservative linear bandits and RL because one will need new constructions of hard environments for different settings. 
We note that new constructions are needed even for non-conservative settings, because simply embedding the hard instances of MAB to other settings \emph{cannot} give the tightest lower bounds. 
See, e.g., Chapter 24 of \cite{lattimore2020bandit} and \cite{DominguesMKV21}.

In this paper, We use a completely different approach.
Our key insights are 1) \textbf{relating the necessary budget to the regret lower bounds of  non-conservative sequential decision-making problems}, 
and 2) obtaining sharp lower bounds in the conservative settings via \textbf{maximizing a quadratic function (cf. Equation~\eqref{eqn:rebuttal6})}.
Comparing with the classical approach, our approach is simpler and more general: ours does not need problem-specific constructions and can automatically transform any lower bound in a non-conservative problem to the corresponding conservative problem.
See Section~\ref{sec:lb} for details.

\footnotetext[2]{Although the lower bound in \citet{wu2016conservative} seems tighter, they require a condition $\frac{\Delta_{0}}{\alpha \mu_{0} + \Delta_{0}} \geq 0.9$.
Under this condition, our lower bound is the same as theirs. Thus ours is more general. See Appendix \ref{app:discussion}.
}

\footnotetext[3]{In \citep{garcelon2020conservative}, the upper bound scales with $r_{b} = \min_{s\in \mathcal{S}, \rho_{0}(s) >0} V_{1}^{\pi_{0}}(s)$ (with $\rho_{0}$ the distribution of the starting state), the minimum of the baseline's value function at the first step over the potential starting states.Here, we assume there is a unique starting state hence $r_{b} = V^{\pi_0}$.}

\subsubsection{Upper Bounds}
\textbf{Improvement over \cite{wu2016conservative} when $\Delta_0$ is known.}
When $\Delta_0$ is known, \citet{wu2016conservative} proposed an algorithm (BudgetFirst) which first plays the baseline policy for enough times and then plays an non-conservative MAB algorithm.
However, their regret bound is not tight because their analysis on the required budget is loose: they accumulate enough budget to play $T$-\textbf{step} exploration where $T$ is the total number of rounds.
Our main technical insight to obtain the tight regret bound is a sharp analysis on the required budget: by relating the minimax regret upper bounds of UCB algorithms, we show the required budget  can be \textbf{independent of $T$}.
See Section~\ref{sec:ub} and~\ref{sec:compare_upper} for details.

\textbf{Sharp upper bounds with unknown $\Delta_0$.}
When  $\Delta_0$ is unknown, 
the paper by \cite{wu2016conservative}, its follow-up papers~\citep{kazerouni2016conservative,garcelon2020improved,zhang2019contextual,garcelon2020conservative}, and our paper, all adopt the same algorithmic template: 1)  build an online estimate on the lower bound performance of each possible exploration policy, and 2) based on the estimated lower bounds, choose an exploration policy or play the baseline policy.


The key difference and the most non-trivial part in different papers is how to analyze $T_0$ (the number of times of executing the baseline policy). 
Existing works upper bound $T_0$ by relating it to the decision criterion for whether to choose the baseline policy or not.
Since for different problem settings, the criteria have different forms, existing papers adopt different problem-specific analyses, and in some settings, the analyses are not tight (e.g., MAB and tabular RL).
Our analysis approach is different from existing ones: we bound $T_0$ via \textbf{maximizing a quadratic function that depends on the minimax regret bounds of non-conservative algorithms and the conservative constraint}.
See Section~\ref{sec:ub} for more details.

%% file: related.tex
Non-conservative exploration has been widely studied in bandits, and minimax optimal algorithms have been provided for the settings considered in this paper~\citep[e.g.][]{lattimore2020bandit}.
The exploration problem has been widely studied also in RL but minimax optimal algorithms have not been provided for all the settings.
For any finite-horizon time-inhomogeneous MDP with $S$ states, $A$ actions and horizon $H$, the minimax regret lower bound is $\Omega(\sqrt{H^3 SAT})$~\citep{DominguesMKV21}, where $T$ denotes the number of episodes.
For any time-inhomogeneous low-rank MDP with $d$-dimensional linear representation, the lower-bound is
$\Omega(\sqrt{d^2H^3T})$~\citep[][Remark 5.8]{Zhou2020nearly}. While several minimax optimal algorithms have been provided for tabular MDPs~\citep[e.g.][]{azar2017minimax,ZanetteB19,zhang2020reinforcement,Zhang0J20,menard2021ucbmq}, the gap between upper bound and lower bound is still open in low-rank MDPs, where LSVI-UCB~\citep{jin2020provably} attains a $\widetilde{O}(\sqrt{d^3H^4T})$,
while ELEANOR~\citep{Zanette2020low} improves to $\widetilde{O}(\sqrt{d^2H^4T})$.

In conservative exploration, previous works focus on designing specific conservative algorithms for different settings.
This conservative scenario was studied in multi-armed bandits~\citep{wu2016conservative}, contextual linear bandits~\citep{kazerouni2016conservative,garcelon2020improved}, contextual combinatorial bandits~\citep{zhang2019contextual} and tabular MDPs~\citep{garcelon2020conservative}.
All these works focused on providing an upper-bound to the regret of a conservative algorithm. 
%
Other problems that have been considered in conservative exploration are combinatorial semi-bandit with exchangeable actions~\citep{Katariya2019interleaving} and contextual combinatorial cascading bandits~\citep{wang2021conservative}.
\citet{du2020onesizefitsall} have recently considered conservative exploration with sample-path constraint.


Our work is also related to safe bandits/RL \citep{AmaniAT19,PacchianoGBJ21,amani2021safe} and constrained RL \citep{altman1999constrained,efroni2020explorationcmdp,DingZBJ20,DingWYWJ21,chen2020efficient}. The setting of safe bandits/RL is different from conservative bandits/RL. Specifically, the safety constraint requires that the expected cost at each stage is below a certain threshold. This constraint is stage-wise, and is independent of the history. On the contrary, the conservative constraint requires that the total reward is not too small. For the constrained MDP, the goal is to maximize the expected reward value subject to a constraint on the expected utility value (value function with respect to another reward function). In conservative RL, however, the agnet aims to maximize the expected reward value subject to the constaint that the (same) reward value is not significantly worse that of the baseline policy.


%% file: preliminaries.tex

The objective of this section is to provide a unified view of the settings considered in this paper, i.e., multi-armed bandits, linear bandits, tabular Markov Decision Processes (MDPs) and low-rank MDPs.
We use the RL formalism which encompasses the bandit settings.

\para{Notations}
We begin by introducing some basic notation. We use $\Delta(\cdot)$ to represent the set of all probability distributions on a set. For $n \in \mathbb{N}_{+}$, we denote $[n]=\{1,2, \ldots, n\}$. We use $O(\cdot), \Theta(\cdot), \Omega(\cdot)$ to denote the big-O, big-Theta, big-Omega notations. We use $\widetilde{O}(\cdot)$ to hide logarithmic factors. We denote $A\gtrsim (\lesssim) B$ if there exists a positive constant $c$ such that $A\geq(\leq) cB$.

\para{Tabular MDPs}
A tabular finite-horizon time-inhomogeneous MDP can be represent as a tuple $M=(\mathcal{S}, \mathcal{A}, H, \{p_h\}_{h=1}^H, s_1, \{r_h\}_{h=1}^H)$, where $\mathcal{S}$ is the state space, $\mathcal{A}$ is the action space, $H$ is the length of each episode and $s_1$ is the initial state.
At each stage $h$, every state-action pair $(s, a)$ is characterized by a reward distribution with mean $r_h(s, a)$ and support in $\left[0, r_{\max }\right]$, and a transition distribution $p_h(\cdot | s, a)$ over next states. We denote by $S=|\mathcal{S}|$ and $A=|\mathcal{A}|$.
A (randomized) policy $\pi \in \Pi$ is a set of functions $\{\pi_h:\mathcal{S}\mapsto \Delta(\mathcal{A})\}_{h\in [H]} $. For each stage $h \in [H]$ and any state-action pair $(s,a)\in \mathcal{S}\times\mathcal{A}$, the value functions of a policy $\pi$ are defined as:
\begin{align*}
    Q^\pi_{h}(s,a)=\mathbb{E}\left[\sum_{h'=h}^H r_{h'}|s_h=s,a_h=a,\pi \right], ~~~
    V_h^{\pi}(s)=\mathbb{E}\left[\sum_{h'=h}^H r_{h'}|s_h=s,\pi \right].
\end{align*}
For each policy $\pi$, we define $V_{H+1}^{\pi}(s)=0$ and $Q_{H+1}^{\pi}(s,a)=0$ for all $s\in\mathcal{S},a\in\mathcal{A}$.
There exists an optimal policy $\pi^\star$ such that $Q^\star_h(s,a)=Q^{\pi_\star}_h(s,a)=\max_\pi Q^\pi_h(s,a)$ satisfy the optimal Bellman equations $Q^\star_h(s,a) = r_h(s,a) + \mathbb{E}_{s' \sim p_h(s,a)}[V^\star_{h+1}(s')]$ and $V^\star_h = \max_{a\in\mathcal{A}} \{Q^\star_h(s,a)\}$. Then the optimal policy is the greedy policy $\pi^\star_h(s) = \arg\max_{a\in\mathcal{A}} \{Q_h^\star(s,a)\}$.

\para{Low-Rank MDPs}
We assume that $\mathcal{S}, \mathcal{A}$ are measurable spaces with possibly infinite number of elements. For algorithmic tractability, we shall restrict the attention to  $\mathcal{A}$ being a finite set with cardinality $A$. When the state space is large or uncountable, value functions cannot be represented in tabular form. A standard approach is to use a parametric representation. Here, we assume that transitions and rewards are linearly representable~\citep{jin2020provably}.
%
\begin{assumption}[Low-rank MDP]\label{asm:lowrank}
An MDP $(\mathcal{S}, \mathcal{A}, H, p, r)$ is a linear MDP with a feature map $\boldsymbol{\phi}: \mathcal{S} \times$
$\mathcal{A} \rightarrow \mathbb{R}^{d}$, if for any $h \in[H]$, there exist $d$ unknown (signed) measures $\boldsymbol{\mu}_{h}=\left(\mu_{h}^{(1)}, \ldots, \mu_{h}^{(d)}\right)$ over $\mathcal{S}$ and an unknown vector $\boldsymbol{\theta}_{h} \in \mathbb{R}^{d}$, such that for any $(x, a) \in \mathcal{S} \times \mathcal{A}$, we have
\begin{equation}
\mathbb{P}_{h}(\cdot \mid x, a)=\left\langle\boldsymbol{\phi}(x, a), \boldsymbol{\mu}_{h}(\cdot)\right\rangle, \quad r_{h}(x, a)=\left\langle\boldsymbol{\phi}(x, a), \boldsymbol{\theta}_{h}\right\rangle .
\end{equation}
Without loss of generality, we assume $\|\boldsymbol{\phi}(x, a)\| \leq 1$ for all $(x, a) \in \mathcal{S} \times \mathcal{A}$, and $\max \left\{\left\|\boldsymbol{\mu}_{h}(\mathcal{S})\right\|,\left\|\boldsymbol{\theta}_{h}\right\|\right\} \leq$
$\sqrt{d}$ for all $h \in[H]$.
\end{assumption}
Under certain technical conditions~\citep[e.g.,][]{shreve1978alternative}, all the properties of tabular MDPs extend to low-rank MDPs.
In addition, the state-action value function of any policy $\pi$ is linearly representable in low-rank MDPs. Formally, for any policy $\pi$ and stage $h \in [H]$, there exists $\theta^\pi_h \in \mathbb{R}^d$ such that $Q_h^\pi(s,a) = \langle \phi(s,a), \theta_h^\pi \rangle$.

\para{Connection between RL and Bandits} To have a unified view, we can represent a multi-armed bandit as a tabular MDP with $S=1$, $A$ actions, $H=1$ and self-loop transitions in $s_1$. In multi-armed bandits, we consider only deterministic policies so that $\Pi = \mathcal{A}$, then $V^\pi(s_1) = r(s_1, \pi(s_1))$ and the optimal policy is simply $\pi^\star = \arg\max_{a\in\mathcal{A}} r(s_1,a)$.
Similarly, a linear bandit can be modeled through low-rank MDPs with $H=1$.
For generality, we allow the action space to be possibly uncounted and we define the value of a deterministic policy $\pi= a$ ($\Pi = \mathcal{A}$) as $V_1^\pi(s_1) = r_1(s_1,a) = \langle \phi(s_1,a), \theta_1 \rangle$. The optimal policy $\pi^\star$ is thus such that $\pi^\star = \arg\max_{a \in \mathcal{A}} \langle \phi(s_1,a), \theta_1 \rangle$. We refer the reader to Appendix~\ref{app:pre} for details. 

%% file: framework.tex

%
With the unified view provided in the previous section, we can consider a generic sequential decision-making problem $\mathfrak{P}$ over $T\in\mathbb{N}^{\star}$ episodes.
We consider the standard online interaction protocol where, at each episode $t\in[T]$, the learning agent $\mathfrak{A}$ selects a policy $\pi_t$, observes and stores a trajectory $(s_i, a_i, r_i)_{i\in[H]}$, updates the policy and restart with the next episode. We evaluate the performance of the learner through the \emph{pseudo-regret}. Let $V^{\pi} = V_1^{\pi}(s_1)$ be the value function of a policy $\pi$, then the regret is defined as:
\begin{equation}\label{eq:regret}
    R_{T}(\mathfrak{P}, \mathfrak{A})=\sum_{t=1}^{T} V^\star -V^{\pi_{t}}.
\end{equation}
In conservative exploration, the learner aims to minimize the regret while guaranteeing that, at any episode $t$, their expected performance is (nearly) above the one of a baseline policy $\pi_0$.
Formally, given a possibly randomized baseline policy $\pi_0 \in \Pi$ and a conservative level $\alpha \in [0,1]$, the learner should satisfy w.h.p.\ that

\begin{equation}\label{eq:conservative.condition}
    \forall t\leq T, \qquad \sum_{j=1}^{t} V^{\pi_{j}} \geq (1-\alpha) \; t \; V^{\pi_{0}}.
\end{equation}
We assume that the value of conservative policy $V^{\pi_{0}}$ is known to the agent. Such assumption can be seen in previous works such as \cite{wu2016conservative,kazerouni2016conservative,garcelon2020improved,garcelon2020conservative}. This assumption is reasonable in practice because usually the baseline policy has been used for a long time and is well-characterized, and its value can be estimated using historical data. Even if we do not know the value of baseline policy, we can estimate it during the algorithm (e.g., Section 3.5 in \citet{wu2016conservative}), and we omit here for simplicity.

\subsection{Budget of a Conservative Algorithm}
Given the set of policies $\{\pi_t\}_{t\in[T]}$ selected by a conservative algorithm $\mathfrak{A}$, we can divide the episodes into the set $\mathcal{T}_{0} = \{ t\leq T \mid \pi_{t} = \pi_{0}\}$ and its complement $\mathcal{T}_{0}^{c} = \{t\leq T\mid \pi_{t} \neq \pi_{0} \} = [T] \setminus \mathcal{T}_{0}$.
The set $\mathcal{T}_{0}^{c}$ denotes the episodes where the algorithm played an exploratory policy, i.e., it had enough budget to satisfy condition~\eqref{eq:conservative.condition} through a policy $\pi_l \neq \pi_0$. This sequence of non-baseline policies $\{\pi_{t}\}_{t\in \mathcal{T}_{0}^{c}}$ defines a new algorithm $\widetilde{\mathfrak{A}}$, that we refer as the non-conservative algorithm. However, the algorithm $\mathfrak{A}$ is conservative therefore, for any $\delta>0$ and $t\in [T]$, we have with probability at least $1-\delta$ that$\sum_{l=1}^{t} V^{\pi_{l}} \geq (1 - \alpha) t V^{\pi_{0}}$. Hence, for any $t\in [T]$ we have:
    \begin{align}\label{eq:temp_def_budget}
        \alpha V^{\pi_{0}} |\mathcal{T}_{0,t}| \geq \sum_{l\in \mathcal{T}_{0,t}^{c}} (1 - \alpha)V^{\pi_{0}} - V^{\pi_{l}},
    \end{align}
    where $\mathcal{T}_{0, t} = \mathcal{T}_{0} \cap [t]$ and $\mathcal{T}_{0, t}^{c} = \mathcal{T}_{0}^{c} \cap [t]$. Taking maximum over $t$ in
    Eq.~\eqref{eq:temp_def_budget}, we have that with high probability the conservative algorithm $\mathfrak{A}$ is such that
    \begin{equation*}
        \alpha V^{\pi_{0}} |\mathcal{T}_{0}|
    \geq 
    \underbrace{
        \max_{t\leq T} \sum_{l\in \mathcal{T}_{0,t}^{c}} (1 - \alpha)V^{\pi_{0}} - V^{\pi_{l}}
    }_{=\mathcal{B}} .
    \end{equation*}
    The quantity on the right of the previous equation is exactly the amount of reward that the conservative algorithm $\mathfrak{A}$ has to collect by playing the baseline policy.
    Hence this quantity acts as a \emph{conservative budget} $\mathcal{B}$. The higher it is, the more $\mathfrak{A}$ needs to play the baseline policy to satisfy the conservative condition. In other words, it is the least amount of reward that an algorithm needs to not violate the conservative constraint.
    We now extend this notion to any (non necessarily conservative) algorithm.

    \begin{definition}\label{def:budget}
        For any $T\in\mathbb{N}^{\star}$, set of episodes $\mathcal{O} \subset [T]$ and arbitrary sequence of policies $\{\pi_{t}\}_{t\in \mathcal{O}}$,
        the budget of this sequence of policies is defined as:
        \begin{align}\label{eq:budget}
            \mathcal{B}_{T}(\mathcal{O}, \{\pi_{t}\}_{t\in \mathcal{O}}) = \max_{t\in \mathcal{O}} \sum_{l\in\mathcal{O} \cap [t]} (1 - \alpha) V^{\pi_{0}} - V^{\pi_{l}}.
        \end{align}
    \end{definition}

%% file: lb.tex
In this section, we leverage the framework introduced in Section~\ref{sec:framework} to build lower bounds for several problems. Our result is based on the notion
of budget defined in Section~\ref{sec:framework}. This notion is used to build an algorithm whose regret
is a lower bound for any conservative algorithm.

\begin{theorem}[Conservative Exploration Regret Lower Bound] \label{thm:lb_general}
    Let's consider a decision-making problem $\mathfrak{P}$ over $T$ steps, a conservative level $\alpha \in[0,1]$, a baseline policy $\pi_{0}$, an algorithm $\mathfrak{A}$ and $\delta\in(0,1)$. We assume that:
    \begin{itemize}[noitemsep,topsep=0pt,parsep=0pt,partopsep=0pt,leftmargin=0.2in]
        \item \emph{Lower-bound for non-conservative exploration.} There exists a $\xi\in\mathbb{R}_{+}$ and $T_{0}\in \mathbb{N}$ such that for any algorithm $\mathfrak{A}'$ there exists an environment (instance of $\mathfrak{P}$) such that with probability at least
        $1 - \delta$, $R_{T}(\mathfrak{P}, \mathfrak{A}') \geq \xi\sqrt{T}$ for $T\geq T_{0}$.
        \item \emph{$\mathfrak{A}$ is conservative.} The algorithm $\mathfrak{A}$ is conservative, that is to say with probability at least $1 - \delta$ for any $t\leq T$, $\sum_{l=1}^{t} V^{\pi_{l}} \geq (1 - \alpha) t V^{\pi_{0}}$.
    \end{itemize}
    Then, there exists an environment (instance of problem $\mathfrak{P}$) and $T_{0}\in \mathbb{N}$ such that with probability at least $1 - \delta$ and $T\geq T_{0}$:
    \begin{align*}
       R_{T}(\mathfrak{A}, \mathfrak{P}) \gtrsim \max\Big\{\xi\sqrt{T}, \frac{\xi^{2}\Delta_{0}}{\alpha V^{\pi_{0}}(\alpha V^{\pi_{0}} + \Delta_{0})}\Big\}.
    \end{align*}
    where $\Delta_0 = V^\star - V^{\pi_0}$ is the sub-optimality gap of policy $\pi_0$.
    \end{theorem}

Theorem~\ref{thm:lb_general} provides a general framework deriving lower-bounds for conservative exploration and highlights the impact of the baseline policy on the regret.
In particular, it shows that in any sequential decision-making problem, after a sufficiently large number of episodes the conservative condition can be verified and the baseline policy has no impact anymore on the learning process.
The only requirement is the knowledge of a lower-bound for the non-conservative case. Before instantiating the result in specific settings, we provide an intuition about how this result is derived and what is the role of the conservative budget $\mathcal{B}$.

    \para{Proof Sketch} Let us consider a conservative algorithm $\mathfrak{A} = \{ \pi_{t} \mid t\leq T\}$, which is associated to a non-conservative algorithm $\widetilde{\mathfrak{A}}
    = \{\pi_{t}\mid t\in \mathcal{T}_{0}^{c} \}$ with $\mathcal{T}_{0}^{c}$ and $\mathcal{T}_{0}$ the set of non-conservative and conservative
    episodes as defined in Sec.~\ref{sec:framework}. Now if $\mathbb{E}\left|\mathcal{T}_{0}\right| \geq \frac{\xi^{2}}{\alpha V^{\pi_{0}} \cdot\left(\alpha V^{\pi_{0}}+\Delta_{0}\right)}$ (i.e. the algorithm plays $\pi_0$  too many times), then the regret caused by $\pi_0$ is at least $\frac{\xi^{2} \Delta_{0}}{\alpha V^{\pi_{0}} \cdot\left(\alpha V^{\pi_{0}}+\Delta_{0}\right)}$. When $\mathbb{E}\left|\mathcal{T}_{0}\right| < \frac{\xi^{2}}{\alpha V^{\pi_{0}} \cdot\left(\alpha V^{\pi_{0}}+\Delta_{0}\right)}$, consider the budget of $\mathcal{T}_0$ defined in Definition \ref{def:budget}:
    \begin{align}\label{eqn:rebuttal6}
                B_{\mathcal{T}_0^c}(\mathcal{A}_{c}) &= \max_{t \in \mathcal{T}_0^c}\mathbb{E} \sum_{k = 1}^{t} [(1 - \alpha) V^{\pi_0} - V^{\pi^t}] =  \max_{t \in \mathcal{T}_0^c} \mathbb{E}[R^{T_0^c}_{\mathfrak{A}}(\mathcal{A}_{c})(t)] - (\alpha V^{\pi_0} + \Delta_0) t,
    \end{align}
    where $\mathbb{E}\left[R_{\mathfrak{A}}^{T_{0}^{c}}\left(\mathcal{A}_{c}\right)(t)\right]$ is the regret incurred by the rounds in $T_{0}^{c}$. Now if $\mathbb{E}\left[R_{\mathfrak{A}}^{T_{0}^{c}}\left(\mathcal{A}_{c}\right)(t)\right] \geq \xi \sqrt{t}$, we have $B_{\mathcal{T}_{0}^{c}}\left(\mathcal{A}_{c}\right)  \gtrsim \frac{\xi^{2}}{\alpha V^{\pi_{0}}+\Delta_{0}}$ by taking maximum on the right handside of \eqref{eqn:rebuttal6} (viewing RHS as a quadratic function of $\sqrt{t}$). Therefore $\mathbb{E}\left|\mathcal{T}_{0}\right| \geq \frac{B_{\mathcal{T}_{0}^{c}}\left(\mathcal{A}_{c}\right)}{\alpha V^{\pi_{0}}} \gtrsim \frac{\xi^{2}}{\alpha V^{\pi_{0}} \cdot\left(\alpha V^{\pi_{0}}+\Delta_{0}\right)}$ and the regret is also no smaller than $\frac{\xi^{2} \Delta_{0}}{\alpha V^{\pi_{0}} \cdot\left(\alpha V^{\pi_{0}}+\Delta_{0}\right)}$, which completes the proof.

    \para{Example of Lower Bounds} 
    For instance, in the multi-armed bandits, by leveraging the lower-bound in~\cite[Thm.~$15.2$][]{lattimore2020bandit}, we can obtain the following corollary of Theorem~\ref{thm:lb_general}.
    This result is more general than the lower bound in \cite{wu2016conservative} where they have a restriction that $\frac{\Delta_{0}}{\alpha \mu_{0} + \Delta_{0}} \geq 0.9$. See Appendix \ref{app:discussion} for details.
    \begin{corollary}\label{cor:lb:mab}
        For any $K\in \mathbb{N}^{\star}$, $\alpha\in [0,1]$, $\mu_{0}\in[0,1]$, $\delta\in(0,1)$ and a conservative algorithm $\mathfrak{A}$ then there exists $\mu\in[0,1]^{K}$
        such that $\sum_{l=1}^{t} \mu_{\pi_{l}} \geq (1 - \alpha)\mu_0 t$ with high probability for any $t\leq T$. Then, for $T \geq \frac{A}{\alpha \mu_0 \cdot (\alpha\mu_0 + \Delta_0)} + \frac{\sqrt{A}}{\alpha \mu_0 + \Delta_0}$,
 $
           R_T(\mu, \mathfrak{A}) \gtrsim \max \Big\{\sqrt{AT},  \frac{A\Delta_0}{\alpha \mu_0 \cdot (\alpha \mu_0 + \Delta_0)}\Big\}.
$
    \end{corollary}

    The generality of Theorem~\ref{thm:lb_general} allows us to derive lower-bounds for conservative exploration in many different problems, where the lower-bound was unknown. Table~\ref{tab:regret.bounds} reports the lower-bound obtained through Theorem~\ref{thm:lb_general}. Please refer to Appendix~\ref{app:nonc.lowerbound} for lower-bounds for non-conservative exploration.
    In linear bandits, 
   the lower bound we obtain matches the result in~\citep{kazerouni2016conservative,garcelon2020improved}, showing the optimality of their algorithms. In tabular MDPs, our result shows
    that the dependence on $S,A$ and $H$ of CUCBVI~\citep{garcelon2020conservative} is not optimal. Finally, by instantiating Theorem~\ref{thm:lb_general} in low-rank MDPs, we obtain the first lower bound for this setting.

%% file: ub.tex
In this section, we show how to leverage the framework of Sec.~\ref{sec:framework} to derive an algorithm
for any conservative sequential decision-making problem. We first show that when knowing $\Delta_{0}$ a simple algorithm achieves a minimax regret,
as prescribed by our lower bound of Sec.~\ref{sec:lb}.
Then, we show how to remove this knowledge without hurting the performance by combining our framework and the idea of lower confidence bound.

\subsection{The \textsf{Budget-Exporation} Algorithm}

    Given a non-conservative algorithm $\widetilde{\mathfrak{A}}$, the minimum amount of rewards needed to play this non-conservative algorithm for $T$ consecutive steps is the budget defined in Def.~\ref{def:budget}.
    Indeed, if we denote by $\{ \tilde{\pi}_{l} \mid l \leq T \}$ the sequence of non-conservative policies executed by $\widetilde{\mathfrak{A}}$, then for any set $\mathcal{O} \subset [T]$ the budget can be rewritten as:
    \begin{align*}
        \mathcal{B}_{T}(\mathcal{O}, \{ \tilde{\pi}_{l} \mid l \leq T \}) &= \max_{t\in \mathcal{O}} \sum_{l\in \mathcal{O}\cap[t]} (1 - \alpha) V^{\pi_{0}} -  V^{\pi_{l}} \\
        &= \max_{t\in \mathcal{O}}
        \sum_{l\in \mathcal{O}\cap[t]}  \Big(
        V^{\star} - V^{\pi_{l}}  - (\Delta_{0} + \alpha V^{\pi_{0}})\big|\mathcal{O}\cap[t]\big|
        \Big) .
    \end{align*}
    Let's define $R_{\mathcal{O}\cap [t]}(\widetilde{\mathfrak{A}}) := \sum_{l\in\mathcal{O}\cap[t]} V^{\star} - V^{\pi_{l}}$ the regret over the
    time steps in $\mathcal{O}$ of the non-conservative algorithm $\widetilde{\mathfrak{A}}$.
    For most non-conservative algorithms with minimax regret bound, $\tilde{R}_{T}(\widetilde{\mathfrak{A}}, \mathcal{O}) = \mathcal{O}(C\sqrt{|\mathcal{O} \cap [t]|})$ w.h.p., where $C\in\mathbb{R}$ is a problem-dependent quantity as in Theorem~\ref{thm:lb_general}. For example, in multi-armed
    bandit $C = \sqrt{A}$ for the \textsf{UCB} algorithm or $C =\sqrt{H^{3}SA}$ for the \textsf{UCBVI-BF} algorithm \citep{azar2017minimax}. This implies that the budget required by $\widetilde{\mathfrak{A}}$ is at least $\frac{C^{2}}{\Delta_{0} + \alpha V^{\pi_{0}}}$. Therefore, the simple algorithm playing the baseline policy for the first $T_{0} := O(\frac{C^2}{(\alpha V^{\pi_{0}}+\Delta_0)\alpha V^{\pi_{0}}})$ steps and then running the non-conservative algorithm $\widetilde{\mathfrak{A}}$, is conservative.
    We call such algorithm \textsf{Budget-Exporation} (see Alg.~\ref{alg:budgetfirst}). This algorithm is conservative and minimax optimal.
    Indeed, we can show (see Theorem~\ref{thm:budgetfirst}) that the regret upper bound of \textsf{Budget-Exporation} matches the lower bounds of Section~\ref{sec:lb}.
    While knowing $\Delta_0$ in advance may be a restrictive assumption, it is interesting that a two-stage algorithm structure (deploying a baseline policy and then a non-conservative policy)
    is enough to achieve minimax optimality.

\setlength{\textfloatsep}{0.2cm}
        \begin{algorithm}[t]\label{alg:budgetfirst}
            \SetAlgoLined
            \KwIn{A non-conservative algorithm $\widetilde{\mathfrak{A}}$, conservative policy cumulative reward $V^{\pi_0}$, conservative level: $\alpha\in (0,1)$
            ,baseline action gap: $\Delta_{0} = V^{\star} - V^{\pi_{0}}$ and a constant $C$}
            Set $B=\frac{C^2}{\alpha V^{\pi_{0}}+\Delta_0}$ and $T_{0}=\frac{B}{\alpha V^{\pi_{0}}}$\;
            \For{$t=1,\ldots,T$}{
                \eIf{$t<T_{0}$}{
                    Play $\pi_0$\;
                }{
                    Play according to $\tilde{\mathfrak{A}}$\;
                }
            }
            \caption{Budget-Exporation}
        \end{algorithm}
%
\begin{theorem}\label{thm:budgetfirst}
    Consider an algorithm $\widetilde{\mathfrak{A}}$, $\delta\in(0,1)$ and constant $C\in\mathbb{R}$  such that with probability at least $1-\delta$, for any $T\geq 1$, $R_T(\widetilde{\mathfrak{A}}) \leq \widetilde{O}(C\sqrt{T})$.
    Then for any $T\geq 1$, the regret of \textsf{Budget-Exporation} is bounded with probability at least $1 - \delta$ by $\widetilde{O}(C\sqrt{T}+\frac{C^2 \Delta_0}{\alpha V^{\pi_{0}}(\alpha V^{\pi_{0}}+\Delta_0)}) $.
\end{theorem}

%

Instantiating Thm.~\ref{thm:lb_general} with $\widetilde{\mathfrak{A}}$ being the UCB algorithm \citep{lattimore2020bandit}, then $C=\sqrt{A}$ and the regret of \textsf{Budget-Exporation} is bounded w.h.p.\ by $\widetilde{O}(\sqrt{AT}+\frac{A \Delta_0}{\alpha V^{\pi_{0}}(\alpha\mu_0+\Delta_0)})$, that matches our novel lower bound introduced in Sec.~\ref{sec:lb}.
Similar results can be obtained for the other settings, see Table~\ref{tab:regret.bounds}.
In linear bandit we consider \textsf{LinUCB} as the non-conservative algorithm, leading to $C = d$.
Similarly, in tabular MDP and low-ran MDPs, we get $C=\sqrt{H^3SA}$ and $C=\sqrt{d^3H^4}$
respectively using
\textsf{UCBVI-BF} \citep{azar2017minimax} and \textsf{LSVI-UCB} \citep{jin2020provably}.
Refer to Table~\ref{tab:regret.bounds} for a complete comparison of the results.


\subsection{The \textsf{LCBCE} Algorithm}

    When $\Delta_0$ is unknown, we aim to use the same idea as \textsf{Budget-Exporation}, that is to say to play a policy different than the baseline one only if the budget is positive.
    To achieve this, we need to build an online estimate of the conservative budget which amounts to build a lower confidence bound (w.h.p.) on the value function
    of any policy $\pi$. Therefore, assuming a non-conservative algorithm $\widetilde{\mathfrak{A}}$ builds such confidence bounds, for example by estimate the MDP as done by~\citet{garcelon2020conservative}, we
    show how our budget framework helps to derive a conservative regret bound.



Let's consider a non-conservative algorithm $\widetilde{\mathfrak{A}} = \{\pi_{t} \mid t\leq T\}$ able to construct a high probability lower bound on the set of selected policies. That is, for any time $t\leq T$ and $\delta\in(0,1)$, $\widetilde{\mathfrak{A}}$ computes a sequence of real
numbers $(\lambda_{t}^{\pi_k}(\delta))_{k\leq t}$ such that with probability at least $1 - \delta$, for all $k\leq t$, $\lambda_{t}^{\pi_{k}}(\delta) \leq V^{\pi_{k}}$. Using these lower bounds, we can define a proxy to the budget for $\widetilde{\mathcal{B}}_{T, \delta}(\mathcal{O}, \widetilde{\mathfrak{A}})$ for any subset $\mathcal{O}\subset[T]$ by
\begin{align}
    \widetilde{B}_{T, \delta}\left(\mathcal{O}, \widetilde{\mathfrak{A}}\right)&=\max _{t \in \mathcal{O}} \sum_{l\in \mathcal{O}\cap[t]}\big((1-\alpha) V^{\pi_0}-\lambda_t^{\pi_l}(\delta)\big),
\end{align}
with $(\pi_{l})_{l\in O}$ the sequence of policies computed by the non-conservative algorithm $\widetilde{\mathfrak{A}}$. Then following from the definition of $(\lambda_{t}^{\pi_{l}}(\delta))_{l\leq t}$, we
have that with probability at least $1 - \delta$ that $\widetilde{B}_{T, \delta}\left(O, \widetilde{\mathfrak{A}}\right) \geq \mathcal{B}_{T}(\mathcal{O}, \widetilde{\mathfrak{A}})$.
This shows that it is possible to compute $\widetilde{B}_{T, \delta}\left(O, \widetilde{\mathfrak{A}}\right)$ without knowledge of the environment and the baseline parameters. The idea
of our algorithm is now to play a non-conservative policy $\pi_{t}$ at time $t$ only if the difference between
the proxy to the budget of $\widetilde{\mathfrak{A}}$ and the reward accumulated by playing the baseline policy is negative. Formally, the condition is $\widetilde{B}_{t, \delta}\left(S_{t} \cup {t}, \widetilde{\mathfrak{A}}\right) \leq \alpha V^{\pi_{0}}(t-1-|S_{t}|)$ where $S_{t}$ is the set of time step where a non-conservative policy was deployed in episodes before $t$.
As a result, the minimum budget that $\widetilde{\mathfrak{A}}$ requires to be conservative is $ \max_{t} \widetilde{B}_{t, \delta}\left(S_{t} \cup {t}, \widetilde{\mathfrak{A}}\right)=\max _{t \in[T]} \sum_{l\in S_{t}}\big( (1-\alpha) V^{\pi_0}-\lambda_t^{\pi_l}(\delta) \big)$.
The algorithm, called \emph{Lower Confidence Bound for Conservative Exploration} (LCBCE),   is detailed in Alg.~\ref{alg:lcb}.

\setlength{\textfloatsep}{0.2cm}
\begin{algorithm}[t]\label{alg:lcb}
    \SetAlgoLined
    \KwIn{A non-conservative algorithm $\widetilde{\mathfrak{A}}$, $\delta\in(0,1)$, lower confidence bounds $\lambda_t^{\pi_k}\leq V^{\pi_k}$, conservative policy value $V^{\pi_0}$, $\alpha\in (0,1)$ }
     Set $B=0$ \tcp*{the accumulated budget}
     Set $t'=0$ \tcp*{the number of steps in which the agent acts as $\widetilde{\mathfrak{A}}$}
     \For{$t=1,2,...,T$}{
      $\widetilde{\mathfrak{A}}$ gives lower bound $\lambda_{t'+1}$ and a policy $\tilde{\pi}_{t'+1}$\;
      Set $\lambda=\sum_{k=1}^{t'}\lambda_{t'+1}^{\tilde{\pi}_k}+\lambda_{t'+1}^{\tilde{\pi}_{t'+1}}$ \tcp*{lower bound of expected total reward}
      \eIf{$\lambda - (t'+1)\alpha V^{\pi_0}<B$}{
       Play $\pi_t=\pi_0$ and set $B=B+\alpha V^{\pi_0}$\;
      }{
       Play $\pi_t=\tilde{\pi}_{t'+1}$ and set  $t'=t'+1$\;
      }
     }
     \caption{Lower Confidence Bound for Conservative Exploration}
\end{algorithm}

Next, we show the regret bound of \textsf{LCBCE}. The proof is in Appendix \ref{app:lower}.

\begin{theorem}\label{thm:unknown_upper}
    Consider an algorithm $\widetilde{\mathfrak{A}}$, $\delta\in(0,1)$ and constant $C\in\mathbb{R}$  such that with probability at least $1-\delta$, for any $T\geq 1$, $R_T(\widetilde{\mathfrak{A}}) \leq \widetilde{O}(C\sqrt{T})$.
    If $\widetilde{\mathfrak{A}}$ computes lower confidence bound such that $\sum_{k=1}^t \big( V^{\pi_k}-\lambda_t^{\pi_k} \big) \leq \widetilde{O}(C\sqrt{T})$ with probability at least $1 - \delta$,
    then for any $T\geq 1$, the regret of \textsf{LCBCE} is bounded with probability at least $1 - \delta$ by $\widetilde{O}(C\sqrt{T}+\frac{C^2 \Delta_0}{\alpha V^{\pi_{0}}(\alpha V^{\pi_{0}}+\Delta_0)}) $.
\end{theorem}
\vspace{-0.2cm}

In the MAB and tabular case, \textsf{LCBCE} paired with \textsf{UCB} achieves a better regret bound compared with previous papers\citep{garcelon2020conservative,wu2016conservative}. We also provide the first minimax optimal bound for the case of unknown baseline parameters.
Finally, in low rank MDPs we recover the same rate as in the case of known baseline.
See Table~\ref{tab:regret.bounds}.

%% file: conclusion.tex
We present a unified framework for conservative exploration in sequential decision-making problems.
This framework can be leveraged to derive both minimax lower and upper bounds. In bandits, we provide novel lower bounds that highlighted the optimality of existing algorithms. In RL, we provide the first lower bound for tabular MDPs and a matching upper bounds, and the first analysis for low rank MDPs.
An interesting question is whether one can leverage this framework to derive problem-dependent logarithmic bounds for conservative exploration. Another direction is to investigate model-free algorithms (e.g., Q-learning~\citep{jin2018q}) for conservative exploration.

%% file: formulation.tex
In this paper we consider conservative bandits and conservative reinforcement learning problems.


\subsection{Conservative multi-armed bandit}
The multi-armed bandit problem is a sequential decision-making task in which a learning agent repeatedly chooses an action (called an arm) and receives a reward corresponding to that action. We assume there are $K+1$ arms, denoted by $\{0, \ldots, K\}$. There is a reward $X_{t, i}$ associated with each arm $i$ at each round $t \in\{1,2, \ldots\}$. In each round $t$, the agent pulls arm $I_{t} \in\{0, \ldots, K\} $ and receives a reward $X_{t, I_{t}}$ corresponding to this arm. The agent does not observe the other rewards $X_{t, j}\left(j \neq I_{t}\right)$.

The learning performance of an agent over a time horizon $T$ is usually measured by its regret, which is the difference between its reward and what it could have achieved by consistently choosing the single best arm in hindsight:

\begin{equation} 
R_{T}=\max _{i \in\{0, \ldots, K\}} \sum_{t=1}^{T} X_{t, i}-X_{t, I_{t}}
\end{equation}

In conservative multi-armed bandits, we assume that the conservative default action is arm 0, and its reward is fixed and is known. That is, $X_{0,t}=\mu_0$ for all $t$. On the other hand, each arm $i>0$ has a stochastic reward $X_{t, i}=\mu_{i}+\eta_{t, i}$, where $\mu_{i} \in[0,1]$ is the expected reward of arm $i$ and $\eta_{t}$ is a random noise such that
\begin{assumption}\label{ass:noise}
Each element $\eta_{t}$ of the noise sequence $\left\{\eta_{t}\right\}_{t=1}^{\infty}$ is conditionally 1-sub-Gaussian, i.e.
\begin{equation}
\forall \zeta \in \mathbb{R}, \quad \mathbb{E}\left[e^{\zeta \eta_{t}} \mid a_{1: t}, \eta_{1: t-1}\right] \leq \exp \left(\frac{\zeta^{2}}{2}\right)
\end{equation}
\end{assumption}
The sub-Gaussian assumption automatically implies that $\mathbb{E}\left[\eta_{t} \mid a_{1: t}, \eta_{1: t-1}\right]=0$ and $\operatorname{Var}\left[\eta_{t} \mid a_{1: t}, \eta_{1: t-1}\right] \leq 1$.

We denote the expected reward of the optimal arm by $\mu^{*}=\max _{i} \mu_{i}$ and the gap between it and the expected reward of the $i$ th arm by $\Delta_{i}=\mu^{*}-\mu_{i}$.

In conservative multi-armed bandits, we constrain the learner to earn at least a $1-\alpha$ fraction of the reward from simply playing arm 0 :
\begin{equation}
\sum_{s=1}^{t} X_{s, I_{s}} \geq(1-\alpha) \sum_{s=1}^{t} X_{s, 0} \quad \text { for all } t \in\{1, \ldots, T\}
\end{equation}
where $\alpha \in(0,1)$ is a predefined constant. The parameter $\alpha$ controls how conservative the agent should be. Small values of $\alpha$ show that only small losses are tolerated, and thus, the agent should be overly conservative, whereas large values of $\alpha$ indicate that the manager is willing to take risk, and thus, the agent can explore more and be less conservative.

\subsection{Conservative Linear Bandits}

In the linear bandit setting, in each round $t$, the agent is given a set of (possibly) infinitely many actions/options $\mathcal{A}$, where each action $a \in \mathcal{A}$ is associated with a feature vector $\phi_{a} \in \mathbb{R}^{d}$. At each round $t$, the agent should select an action $a_{t} \in \mathcal{A} .$ Upon selecting $a_{t}$, the agent observes a random reward $X_{t}$ generated as

\begin{equation}
X_{t, a_{t}}=\left\langle\theta^{*}, \phi_{a_{t}}\right\rangle+\eta_{t},
\end{equation}

where $\theta^{*} \in \mathbb{R}^{d}$ is the unknown reward parameter, $\left\langle\theta^{*}, \phi_{a_{t}}\right\rangle=r_{a_{t}}$ is the expected reward of action $a_{t}$ at time $t$, i.e., $r_{a_{t}}=\mathbb{E}\left[X_{t, a_{t}}\right]$, and $\eta_{t}$ is a random noise that satisfies Assumption \ref{ass:noise}.

We also make the following standard assumption on the unknown parameter $\theta^{*}$ and feature vectors:
\begin{assumption}
There exist constants $B, D \geq 0$ such that $\left\|\theta^{*}\right\|_{2} \leq B,\left\|\phi_{a}\right\|_{2} \leq D$, and $\left\langle\theta^{*}, \phi_{a}\right\rangle \in[0,1]$, for all $t$ and all $a \in \mathcal{A} .$
\end{assumption}

We define $\mathcal{B}=\left\{\theta \in \mathbb{R}^{d}:\|\theta\|_{2} \leq B\right\}$ and $\mathcal{F}=\left\{\phi \in \mathbb{R}^{d}:\|\phi\|_{2} \leq D,\left\langle\theta^{*}, \phi\right\rangle \in[0,1]\right\}$ to be
the parameter space and feature space, respectively.

Similar to multi-armed bandits, the goal of the agent is to minimize the following regret:
\begin{equation} \label{def:regret:mab}
R_{T}=\max _{a \in \mathcal{A}} \sum_{t=1}^{T} X_{t, a}-X_{t, a_{t}}
\end{equation}
which is the difference between the cumulative reward of the optimal action and agent’s strategies.

In the conservative linear bandit setting, at each round $t$, there exists a conservative action $b \in \mathcal{A}_{t}$ and selecting $b$ incurs expected reward $r_{b}$. We assume that $r_{b}$ is known, and the conservative action is not relevant to the underlying parameter $\theta_{*}$. We constrain the learner to earn at least a $1-\alpha$ fraction of the reward from simply playing arm $b$:
\begin{equation}
\sum_{i=1}^{t} r_{a_{i}} \geq(1-\alpha) \sum_{i=1}^{t} r_{b}, \quad \forall t \in [T]
\end{equation}

\subsection{Conservative Tabular MDPs}
We consider conservative exploration in finite horizon tabular MDPs. An MDP can be represent as $M=(\mathcal{S}, \mathcal{A}, H, p, r)$, where $\mathcal{S}$ is the state space, $\mathcal{A}$ is the action space, $H$ is the length of each episode. Every state-action pair $(s, a)$ is characterized by a reward distribution with mean $r(s, a)$ and support in $\left[0, r_{\max }\right]$, and a transition distribution $p(\cdot \mid s, a)$ over next states. We denote by $S=|\mathcal{S}|$ and $A=|\mathcal{A}|$.  In each episode, the agent starts from an initial state $s_{1}$. At each step $h \in[H]$, the agent takes action $a_{h}$ in state $s_{h}$ and receive a random reward $r_{h}$ with mean $r(s,a)$, and transits to state $s_{h+1}$ according to the distribution $p(\cdot \mid s, a)$.

A (randomized) policy $\pi$ is a set of functions $\{\pi_h:\mathcal{S}\mapsto \Delta(\mathcal{A})\}_{h\in [H]} $. Given a policy $\pi$, a level $h\in[H]$ and a state-action pair $(s,a)\in \mathcal{S}\times\mathcal{A}$, the $Q$ function and the value function are defined as:

\begin{align*}
    Q^\pi_{h}(s,a)&=\mathbb{E}[\sum_{h'=h}^H r_{h'}|s_h=s,a_h=a,\pi],\\
    V_h^{\pi}(s)&=\mathbb{E}[\sum_{h'=h}^H r_{h'}|s_h=s,\pi].
\end{align*}

We let $V_{H+1}(s)=0$ and $Q_{H+1}(s,a)=0$ for all $s\in\mathcal{S},a\in\mathcal{A}$. We use $Q^*_h$ and $V^*_h$ to denote the optimal $Q$-function and $V$-function at level $h\in[H]$ without corruptions, which satisfies $Q^*_h(s,a)=\max_\pi Q^\pi_h(s,a)$ and $V^*_h(s)=\max_a Q^*(s,a)$ respectively.

In conservative tabular MDPs, at the beginning of each episode $t$, the agent can choose to run a conservative policy $\pi_0$, which will give the agent a fixed reward $V_1^{\pi_0}$ and ends the episode immediately, or choose to explore in the target MDP $M$ with policy $\pi_k$, and will receive a total reward $V_1^{\pi_t}$. Our goal is to minimize the following regret
\begin{equation}
R_T=\sum_{t=1}^{T} V_{1}^{*}\left(s_{1}\right)-V_{1}^{\pi_{t}}\left(s_{1}\right)
\end{equation}
while satisfying the following conservative constraint
\begin{equation}
\sum_{j=1}^{t} V_{1}^{\pi_{j}}(s_{1}) \geq (1-\alpha) t V_{1}^{\pi_{0}}(s_{1}), \quad \forall t \in [T].
\end{equation}

\subsection{Conservative Linear MDPs}

The conservative linear MDP setting is nearly the same as tabular MDPs, except that $\mathcal{S}$ is a measurable space with possibly infinite number of elements and $\mathcal{A}$ is a finite set with cardinality $A$. We assume that the transition kernels and the reward function are assumed to be linear \citep{jin2020provably}.

\begin{assumption}[Linear MDP]
An MDP $(\mathcal{S}, \mathcal{A}, H, p, r)$ is a linear MDP with a feature map $\boldsymbol{\phi}: \mathcal{S} \times$
$\mathcal{A} \rightarrow \mathbb{R}^{d}$, if for any $h \in[H]$, there exist $d$ unknown (signed) measures $\boldsymbol{\mu}_{h}=\left(\mu_{h}^{(1)}, \ldots, \mu_{h}^{(d)}\right)$ over $\mathcal{S}$ and an unknown vector $\boldsymbol{\theta}_{h} \in \mathbb{R}^{d}$, such that for any $(x, a) \in \mathcal{S} \times \mathcal{A}$, we have
\begin{equation}
\mathbb{P}_{h}(\cdot \mid x, a)=\left\langle\boldsymbol{\phi}(x, a), \boldsymbol{\mu}_{h}(\cdot)\right\rangle, \quad r_{h}(x, a)=\left\langle\boldsymbol{\phi}(x, a), \boldsymbol{\theta}_{h}\right\rangle .
\end{equation}
Without loss of generality, we assume $\|\boldsymbol{\phi}(x, a)\| \leq 1$ for all $(x, a) \in \mathcal{S} \times \mathcal{A}$, and $\max \left\{\left\|\boldsymbol{\mu}_{h}(\mathcal{S})\right\|,\left\|\boldsymbol{\theta}_{h}\right\|\right\} \leq$
$\sqrt{d}$ for all $h \in[H]$.
\end{assumption}

%% file: appendix.tex
\section{Lower Bounds for Non-Conservative Exploration}\label{app:nonc.lowerbound}

\begin{lemma}[Lower Bound for Multi-Armed Bandit] \label{lemma:lb:mab}
    Let $K > 1$ and $T \geq  k - 1$. Then for any multi-armed bandit algorithm, there exists a mean vector $\mu \in [0, 1]^K$ such that 
    \begin{align*}
        \mathbb{E}[R_T] \gtrsim \sqrt{KT}.
    \end{align*}
\end{lemma}

\begin{proof}
    See Theorem 15.2 of \citet{lattimore2020bandit} for a detailed proof.
\end{proof}

\begin{lemma}[Lower Bound for Linear Bandit] \label{lemma:lb:lb}
    Let $d \le 2T$. Then for any linear bandit algorithm, there exists a parameter $\theta \in \mathbb{R}^d$ such that 
    \begin{align*}
        \mathbb{E}[R_T] \gtrsim d\sqrt{T}.
    \end{align*}
\end{lemma}

\begin{proof}
    See Theorem 24.2 of \citet{lattimore2020bandit} for a detailed proof.
\end{proof}

\begin{lemma}[Lower Bound for Tabular RL] \label{lemma:lb:tabular}
   Let $T \geq SA$. Then for any bandit RL algorithm, there exists an MDP such that 
   \begin{align*}
       \mathbb{E}[R_T] \gtrsim \sqrt{SAH^3T}.
   \end{align*}
\end{lemma}

\begin{proof}
    See \citet{jaksch2010near,azar2017minimax,jin2018q} for a detailed proof.
\end{proof}

\begin{lemma}[Lower Bound for Linear MDP] \label{lemma:lb:linmdp}
   Let $T \geq d$. Then for any bandit RL algorithm, there exists an MDP such that 
   \begin{align*}
       \mathbb{E}[R_T] \gtrsim \sqrt{d^2H^3T}.
   \end{align*}
\end{lemma}

\begin{proof}
    This lower bound is obtained by extrapolating the lower bounds of linear bandit and tabular RL. 
\end{proof}

\section{Detailed Proof for Lower Bounds} \label{app:upper}

\input{app_proof_lower_bound.tex}

\section{Detailed Proof for Upper Bounds} \label{app:lower}
\subsection{Proof of Theorem \ref{thm:budgetfirst}}
\begin{proof}
 Given a non-conservative algorithm $\widetilde{\mathfrak{A}}$, the minimum amount of rewards needed to play this non-conservative algorithm for $T$ consecutive steps is the budget defined in Def.~\ref{def:budget}.
    Indeed, if we denote by $\{ \tilde{\pi}_{l} \mid l \leq T \}$ the sequence of non-conservative policies executed by $\widetilde{\mathfrak{A}}$, then for any set $\mathcal{O} \subset [T]$ the budget can be rewritten as:
    \begin{align*}
        \mathcal{B}_{T}(\mathcal{O}, \{ \tilde{\pi}_{l} \mid l \leq T \}) &= \max_{t\in \mathcal{O}} \sum_{l\in \mathcal{O}\cap[t]} (1 - \alpha) V^{\pi_{0}} -  V^{\pi_{l}} \\
        &= \max_{t\in \mathcal{O}}
        \sum_{l\in \mathcal{O}\cap[t]}  \Big(
        V^{\star} - V^{\pi_{l}}  - (\Delta_{0} + \alpha V^{\pi_{0}})\big|\mathcal{O}\cap[t]\big|
        \Big) .
    \end{align*}
    Let's define $R_{\mathcal{O}\cap [t]}(\widetilde{\mathfrak{A}}) := \sum_{l\in\mathcal{O}\cap[t]} V^{\star} - V^{\pi_{l}}$ the regret over the
    time steps in $\mathcal{O}$ of the non-conservative algorithm $\widetilde{\mathfrak{A}}$.
    Since $R_{t}(\widetilde{\mathfrak{A}}, \mathcal{O}) = \mathcal{O}(C\sqrt{|\mathcal{O} \cap [t]|})$ w.h.p., where $C\in\mathbb{R}$ is a problem-dependent quantity as in Theorem~\ref{thm:lb_general}. Therefore, we have
    \begin{align*}
        \mathcal{B}_{T}(\mathcal{O}, \{ \tilde{\pi}_{l} \mid l \leq T \}) &= \max_{t\in \mathcal{O}} \sum_{l\in \mathcal{O}\cap[t]} (1 - \alpha) V^{\pi_{0}} -  V^{\pi_{l}} \\
        &= \max_{t\in \mathcal{O}}
        \sum_{l\in \mathcal{O}\cap[t]}  \Big(
        \mathcal{O}(C\sqrt{|\mathcal{O} \cap [t]|})  - (\Delta_{0} + \alpha V^{\pi_{0}})\big|\mathcal{O}\cap[t]\big|
        \Big) .
    \end{align*}
    
    Let $f(x)=C\sqrt{x}-(\Delta_{0} + \alpha V^{\pi_{0}})x$, then we have $f(x)\leq \frac{C^{2}}{\Delta_{0} + \alpha V^{\pi_{0}}}$ This implies that the budget required by $\widetilde{\mathfrak{A}}$ is at least $\frac{C^{2}}{\Delta_{0} + \alpha V^{\pi_{0}}}$. Therefore, the simple algorithm playing the baseline policy for the first $t_{0} := O(\frac{\xi}{\alpha V^{\pi_{0}}+\Delta_0})$ steps and then running the non-conservative algorithm $\widetilde{\mathfrak{A}}$, is conservative. This is actually the algorithm \textsf{BudgetFirst}. 
    
    The regret of \textsf{BudgetFirst} can be bounded as
    \begin{align*}
        Reg(T)\leq t_{0}+R_{t}(\widetilde{\mathfrak{A}}, \mathcal{O}) = \mathcal{O}(\frac{\xi}{\alpha V^{\pi_{0}}+\Delta_0}+C\sqrt{|\mathcal{O} \cap [t]|})
    \end{align*}
    Thus we finish the proof.
\end{proof}
Now we discuss the regret upper bound for different setups. For multi-armed bandit, the UCB algorithm \citep{lattimore2020bandit} gives us the following guarantee.
\begin{lemma}[Upper Bound for Multi-Armed Bandit]
The regret of UCB can be upper bounded by
\begin{equation}
R_{T} \leq 8 \sqrt{T k \log (T)}+3 \sum_{i=1}^{k} \Delta_{i}
\end{equation}
\end{lemma}
\begin{proof}
See Theorem 7.2 in \citet{lattimore2020bandit} for details.
\end{proof}

For linear bandits, the  LinUCB algorithm \citep{lattimore2020bandit} gives us the following guarantee.
\begin{lemma}[Upper Bound for Linear Bandit]
The regret of LinUCB can be upper bounded by
\begin{equation}
R_{T} \leq C d \sqrt{T} \log (T L)
\end{equation}
where $C>0$ is a suitably large universal constant.
\end{lemma}
\begin{proof}
See Corollary 19.3 in \citet{lattimore2020bandit} for details.
\end{proof}

For tabular RL, the UCBVI-BF algorithm in \citet{azar2017minimax} gives us the following guarantee.
\begin{lemma}[Upper Bound for Tabular RL]
The regret of UCBVI-BF can be upper bounded by
\begin{equation}
R_{T} \leq O(\sqrt{H^3SAT})
\end{equation}
\end{lemma}
\begin{proof}
See \citet{azar2017minimax} for details.
\end{proof}

For linear MDP, the LSVI-UCB algorithm in \citet{jin2020provably} gives us the following guarantee.
\begin{lemma}[Upper Bound for Linear MDP]
the total regret of LSVI-UCB is upper bounded by 
\begin{equation}
R_{T} \leq \tilde{\mathcal{O}}\left(\sqrt{d^{3} H^{4} T }\right).
\end{equation}
\end{lemma}
\begin{proof}
See \cite{jin2020provably} for details.
\end{proof}

\subsection{Proof of Theorem \ref{thm:unknown_upper}}

\begin{proof}
Given an LCB algorithm $\tilde{\mathfrak{A}}$, suppose it maintains lower confidence bound $\lambda_t^{\pi_k}(\delta)\leq V^{\pi_k}$ with probability at least $1-\delta$ that satisfies $\sum^t_{k=1}(V^{\pi_k}-\lambda_t^{\pi_k})\leq \tilde{O}(C\sqrt{t})$. Let $S_t$ to be the set of time step where a non-conservative policy was deployed in episodes before $t$. The additional budget needed by the algorithm can be written as:
\begin{align*}
    \tilde{\mathcal{B}}_{T}(S_T, \mathfrak{A}) &= \max_{t\in [T]} \sum_{l\in S_t} [(1 - \alpha) V^{\pi_{0}} -  \lambda_t^{\pi_{l}}(\delta)] \\
        &= \max_{t\in [T]}
        \sum_{l\in S_t}  \Big(
        V^{\star} - V^{\pi_{l}} + V^{\pi_{l}} - \lambda_t^{\pi_l}(\delta) \Big) - (\Delta_{0} + \alpha V^{\pi_{0}})\big|S_t\big|
        \\
        &\leq  \max_{t\in [T]}
        \sum_{l\in S_t}  \Big(
        V^{\star} - V^{\pi_{l}}\Big) + \tilde{O}(C\sqrt{|S_t|})- (\Delta_{0} + \alpha V^{\pi_{0}})\big|S_t\big|
        \\
        &= \max_{t\in [T]}  R_{S_t}(\tilde{\mathfrak{A}})+\tilde{O}(C\sqrt{|S_t|})- (\Delta_{0} + \alpha V^{\pi_{0}})\big|S_t\big|
\end{align*}
Note that $R_{S_t}(\tilde{\mathfrak{A}})\leq \tilde{O}(C\sqrt{|S_t|})$, so the last line can be upper bounded by $\max_{t\in [T]} \Big(\tilde{O}(C\sqrt{|S_t|})- (\Delta_{0} + \alpha V^{\pi_{0}})\big|S_t\big|
        \Big)$. This is a quadratic function $g(x)=\tilde{O}(C\sqrt{x})- (\Delta_{0} + \alpha V^{\pi_{0}})x$ with variable $x=\sqrt{|S_t|}$, we have $g(x)\leq \tilde{O}(\frac{C^2}{\Delta_0+\alpha V^{\pi_0}})$ as a result. In other words, we show that with high probability, \textsf{LCBCE} only need to accumulate $\tilde{\mathcal{B}}_{T}(S_T, \mathfrak{A})\leq \tilde{O}(\frac{C^2}{\Delta_0+\alpha V^{\pi_0}})$. Since playing the baseline policy yields $\alpha V^{\pi_0}$ budget, \textsf{LCBCE} play the baseline policy for at most $\tilde{O}(\frac{C^2}{\alpha V^{\pi_0} (\Delta_0+\alpha V^{\pi_0})})$ times. Hence, the total regret incurred can be written as:
        \begin{align*}
            R_T(\mathfrak{A})=R_{S_T}(\tilde{\mathfrak{A}})+\tilde{O}(\frac{C^2\Delta_0}{\alpha V^{\pi_0} (\Delta_0+\alpha V^{\pi_0})})\leq \tilde{O}(C\sqrt{T}+\frac{C^2\Delta_0}{\alpha V^{\pi_0} (\Delta_0+\alpha V^{\pi_0})})
        \end{align*}
    Thus we finish the proof.
\end{proof}

Proof of Corollaries of Theorem~\ref{thm:unknown_upper}
Below we discuss the lower confidence bound for different setups. 
\paragraph{Multi-armed Bandits}
For the MAB setting, we can calculate the lower confidence bound simultaneously with the upper confidence bound as 
\begin{equation}
    \max \left\{0, \hat{\mu}_{i}(t-1)-\sqrt{\psi^{\delta}\left(T_{i}(t-1)\right) / T_{i}(t-1)}\right\}
\end{equation}
where $\psi^{\delta}(s)=2 \log \left(K s^{3} / \delta\right)$ and $T_{i}(t-1)$ is the times agent pulls arm $i$ until time $t-1$. $\hat{\mu}_{i}(t-1)$ is the empirical reward. This is similar to the calculation of UCB in \citet{lattimore2020bandit}.

\paragraph{Linear Bandits}
For the linear bandit setting, the lower confidence bound can be chosen as follows: first, we calculate the optimal action
\begin{equation}
    \left(a_{t}^{\prime}, \widetilde{\theta}_{t}\right) \in \arg \max _{(a, \theta) \in \mathcal{A}_{t} \times \mathcal{C}_{t}}\left\langle\theta, \phi_{a}^{t}\right\rangle
\end{equation}
where $\mathcal{C}_{t+1}$ is the confidence set $\mathcal{C}_{t+1}=\left\{\theta \in \mathbb{R}^{d}:\left\|\theta-\widehat{\theta}_{t}\right\|_{V_{t}} \leq \beta_{t+1}\right\}$. Then, we calculate $L_{t}=\min _{\theta \in \mathcal{C}_{t}}\left\langle\theta, z_{t-1}+\phi_{a_{t}^{\prime}}\right\rangle$, where $z_{t-1}=\sum_{i=1}^{t-1} \phi_{a_{i}} $. Then $L_t$ is a lower confidence bound of action $a_t^{\prime}$.

\paragraph{Tabular MDP}
For tabular MDP setting, the upper bound of the $Q$ function can be calculated as $Q_h(s,a)=r(s,a)+\hat{P}_hV_{h+1}(x,a)+b_h(s,a)$, where the bonus function is chosen to be $b_h=\tilde{O}(\sqrt{\frac{\mathrm{Var}(V_{h+1})}{N(s,a)}}+\frac{H}{N(s,a)})$ in \citet{azar2017minimax}. To obtain a high probability lower confidence bound, we substitute $b_h(s,a)$ with $-b_h(s,a)$. We use $Q^l_h$ and $V^l_h$ to denote the lower bound of $Q_h$ and $V_h$ respectively,
\begin{align*}
        &V^l_{h+1}(\cdot)=\max_a Q^l_{h+1}(\cdot,a)\\
    &Q^l_h(\cdot,\cdot)=r(\cdot,\cdot)+\hat{P}_hV^l_{h+1}(\cdot,\cdot)-b_h(\cdot,\cdot),
\end{align*}
then $V^l_h$ is a lower confidence bound of $V_h$ with high probability.

\paragraph{Linear MDP}
For linear MDP setting, the lower confidence bound can be obtained by reversing the sign of the bonus term of the upper confidence bound in \citet{jin2020provably}: 
\begin{align*}
    &\Lambda_h\leftarrow \sum^{k-1}_{\tau=1}\phi(x^\tau_h,a^\tau_h)\phi(x^\tau_h,a^\tau_h)^T+\lambda \mathbf{I}\\
    &w_h\leftarrow \Lambda_h^{-1}\sum^{k-1}_{\tau=1}\phi(x^\tau_h,a^\tau_h)[r_h(x^\tau_h,a^\tau_h)+\max_a Q_{h+1}(x^\tau_{h+1},a)]\\
    &Q_h(\cdot,\cdot)\leftarrow\max\{w_h^T \phi(\cdot,\cdot)-\beta[\phi(\cdot,\cdot)^T\Lambda_h^{-1}\phi(\cdot,\cdot)]^{1/2},0\}\\
    &V_h(\cdot)\leftarrow \max_a Q_h(\cdot,a)
\end{align*}


We note that for all these settings, we have $\sum^t_{k=1}(V^{\pi_k}-\lambda_t^{\pi_k})\leq \tilde{O}(C\sqrt{t})$ with corresponding problem-dependent constant $C$. An easy way to see this is to use symmetry. For the above LCB algorithms, we reverse the sign of the bonus term of the upper confidence bound to obtain lower confidence bound. For example in the tabular MDP case, the regret can be bounded by 
\begin{align*}
R_T\leq \sum_{k=1}^K V^u_{k,1}-V^{\pi_k}\leq \tilde{O}(\sum_{k=1}^K\sum_{h=1}^H b_{k,h})\leq \tilde{O}(C\sqrt{T}).
\end{align*}
Using the fact that $\sum_{k=1}^K V^u_{k,1}-V^l_{k,1}=O(\sum_{k=1}^K\sum_{h=1}^H b_{k,h})$, we have $\sum_{k=1}^K V^{\pi_k}-V^l_{k,1}\leq \tilde{O}(\sum_{k=1}^K\sum_{h=1}^H b_{k,h})$. therefore we can deduce that $\sum_{k=1}^K V^{\pi_k}-V^l_{k,1}\leq \tilde{O}(C\sqrt{T})$.

Using the same techniques, we can prove this property for the other settings.

\section{Comparison with \citet{wu2016conservative}'s Lower Bound}
\label{app:discussion}
First, we restate the lower bound of \citet{wu2016conservative} below.
\begin{theorem}[Restatement of Theorem 9 in \citet{wu2016conservative}]
Suppose for any $\mu_{i} \in[0,1](i>0)$ and $V^{\pi_0}$ satisfying
$$
\min \left\{V^{\pi_0}, 1-V^{\pi_0}\right\} \geq \max \{1 / 2 \sqrt{\alpha}, \sqrt{e+1 / 2}\} \sqrt{K / T},
$$
an algorithm satisfies $\mathbb{E}_{\mu} \sum_{t=1}^{T} X_{t, I_{t}} \geq(1-\alpha) V^{\pi_0}$ T. Then there is some $\mu \in[0,1]^{K}$ such that its expected regret satisfies $\mathbb{E}_{\mu} R_{n} \geq B$ where
$$
B=\max \left\{\frac{K}{(16 e+8) \alpha V^{\pi_0}}, \frac{\sqrt{K T}}{\sqrt{16 e+8}}\right\}.
$$
\end{theorem}

Here $V^{\pi_0}$ is the reward of the conservative policy, $K$ is the number of arms, $T$ is the number of episodes. Compared with our result, the main difference is in the first term, where we have an additional coefficient $\frac{\Delta_{0}}{\alpha V^{\pi_0}+\Delta_{0}}$, which makes our result seems worse. However, as we will show below, in the hard instance of the proof in \citet{wu2016conservative}, $\frac{\Delta_{0}}{\alpha V^{\pi_0}+\Delta_{0}}$ is \textbf{lower bounded by an absolute constant}. Therefore, our lower bound actually implies the result of \citet{wu2016conservative}.

When proving the first term $\frac{K}{(16 e+8) \alpha V^{\pi_0}}$ in the lower bound, \citet{wu2016conservative} requires that the parameters should satisfy the following conditions (see Case 2 in their proof):
\begin{gather*}
    \alpha<\frac{\sqrt{K}}{V^{\pi_0} \sqrt{(16 e+8) T}}, \qquad
    \Delta_0=\frac{K}{4 \alpha V^{\pi_0} T}.
\end{gather*}
With these conditions we immediately have 
\begin{equation*}
    \frac{\alpha V^{\pi_0}}{\Delta_0}=\frac{4\alpha^2 (V^{\pi_0})^2 T}{K}<\frac{4}{16e+8},
\end{equation*}
which implies 
$$1>\frac{\Delta_{0}}{\alpha V^{\pi_0}+\Delta_{0}}> \frac{1}{\frac{4}{16e+8}+1} >0.9. $$
Therefore, this factor only has a constant effect, and we can recover the result of \citet{wu2016conservative}.

\section{Comparison with \cite{wu2016conservative}'s Upper Bound with Known $\Delta_0$}
\label{sec:compare_upper}

Here we discuss why the regret bound of BudgetFirst algorithm in \cite{wu2016conservative} is not tight and why our analysis improves theirs.
In BudgetFirst, they require the number of times the agent plays $\pi_0$ to satisfy
\begin{equation}
\left(\forall t_{0} \leq t \leq T\right) \quad t V^{\pi_0}-R_{\text {worst }} \geq(1-\alpha) T V^{\pi_0}
\end{equation}
where $R_{\text {worst }}=O\left(\sqrt{K T \log \left(\frac{\log (T)}{\delta}\right)}\right)$ is the worst case regret of the non-conservative algorithm \textbf{in $T$ steps}. In other words, they accumulate budget by playing $\pi_0$ so that the budget can compensate for the \textbf{$T$-step} exploration of the non-conservative algorithm.

However, it is not necessary to have this much budget. Let us look at the analysis in our algorithm. In our Budget-Exploration, the budget needed can be written as
\begin{align*}
        \mathcal{B}_{T}(\mathcal{O}, \{ \tilde{\pi}_{l} \mid l \leq T \}) &= \max_{t\in \mathcal{O}} \sum_{l\in \mathcal{O}\cap[t]} (1 - \alpha) V^{\pi_{0}} -  V^{\pi_{l}} \\
        &= \max_{t\in \mathcal{O}}
        \sum_{l\in \mathcal{O}\cap[t]}  \Big(
        V^{\star} - V^{\pi_{l}}\Big)  - (\Delta_{0} + \alpha V^{\pi_{0}})\big|\mathcal{O}\cap[t]\big|
         .
\end{align*}
Let us define $R_{\mathcal{O}\cap [t]}(\widetilde{\mathfrak{A}}) := \sum_{l\in\mathcal{O}\cap[t]} V^{\star} - V^{\pi_{l}}$ the regret over the time steps in $\mathcal{O}$ of the non-conservative algorithm $\widetilde{\mathfrak{A}}$.
For UCB algorithm in MAB, $\tilde{R}_{t}(\widetilde{\mathfrak{A}}, \mathcal{O}) = \mathcal{O}(\sqrt{K|\mathcal{O} \cap [t]|})$ w.h.p., where $K\in\mathbb{R}$ is the number of arms.

Now $$\mathcal{B}_{T}(\mathcal{O}, \{ \tilde{\pi}_{l} \mid l \leq T \})=\max_{t\in \mathcal{O}} R_{\mathcal{O}\cap [t]}(\widetilde{\mathfrak{A}})-  (\Delta_{0} + \alpha V^{\pi_{0}})\big|\mathcal{O}\cap[t]\big|$$
\textbf{Note that the RHS is maximized when $\big|\mathcal{O}\cap[t]\big|= O(\frac{(\Delta_{0} + \alpha V^{\pi_{0}})^2}{K})$, but not when $\big|\mathcal{O}\cap[t]\big|= T$.} This means that we do not need to consider the $T$-step regret as in \citep{wu2016conservative}, which is an over-conservative estimate.

%% file: app_proof_lower_bound.tex

\begin{proof}[Proof of Theorem \ref{thm:lb_general}]
  { Let's consider any sequential decision making problem $\mathfrak{A}$ (for instance  a multi-armed bandit problem, linear bandit, tabular RL or linear RL) such that
  there exists $\xi\in\mathbb{R}$ (a constant solely depending on the sequential decision making problem, e.g., the dimension in linear problems or the number of action in tabular problems), an instance of problem $\mathfrak{A}$ where for a number of time steps $T$ large enough and any algorithm $\mathcal{A}$ we have that:
   \begin{align} \label{eq:200}
    \mathbb{E}[R^{T}_{\mathfrak{A}}(\mathcal{A})] \geq  \xi \sqrt{T},
   \end{align}
   with $R^{T}_{\mathfrak{A}}(\mathcal{A})$ the regret of algorithm $\mathcal{A}$ in problem $\mathfrak{A}$.
   For instance, in the MAB case $\xi = \sqrt{K-1}/27$ with $K$ the number of arms.}
    { Using this non-conservative lower bound, we show our lower bound for the conservative setting for the problem
    $\mathfrak{A}$ with a baseline policy $\pi_{0}$. To do so, let's consider any conservative algorithm (that is to say it satisfies Eq. \eqref{eq:conservative.condition})
    noted as $\mathcal{A}_{c}$. We assume this algorithms selects policies $(\pi^{t})_{t\in[T]}$ and let $\mathcal{T}_0$ denotes the set of rounds in $\{1, \dots, T\}$ where $\mathcal{A}_{c}$ selects the conservative policy $\pi_0$. Here $T \ge \frac{\xi^2}{\alpha V^{\pi_0} \cdot (\alpha V_1^{\pi_0} + \Delta_0)} + \frac{\xi^2}{4(\alpha V^{\pi_0} + \Delta_0)^2}$.

    We now distinguish two cases:

        \begin{itemize}
            \item If $\mathbb{E} |\mathcal{T}_0| \geq \frac{\xi^2}{\alpha V^{\pi_0} \cdot (\alpha V^{\pi_0} + \Delta_0)}$, then the definition of the regret implies that:
            \begin{align} \label{eq:201}
                \mathbb{E}[R^{T}_{\mathfrak{A}}(\mathcal{A}_{c})] \geq \mathbb{E} \sum_{t \in \mathcal{T}_0} [V^{*} - V^{\pi^t} ] = \mathbb{E} |\mathcal{T}_0| \cdot \Delta_0 \geq \frac{\xi^2\Delta_0}{\alpha V^{\pi_0} \cdot (\alpha V^{\pi_0} + \Delta_0)}.
            \end{align}
            \item If $\mathbb{E}|\mathcal{T}_0| < \frac{\xi^2}{\alpha V^{\pi_0} \cdot (\alpha V^{\pi_0} + \Delta_0)}$, then let's note $\mathcal{T}_0^c = \{i_1, i_2, \cdots, i_{|\mathcal{T}_0^c|}\}$ the set of time steps where $\mathcal{A}_{c}$ does not execute
            the conservative policy $\pi_{0}$. Considering the budget as we have defined in Def.~\ref{def:budget} we have:
            \begin{align}\label{eq:202}
                B_{\mathcal{T}_0^c}(\mathcal{A}_{c}) &= \max_{t \in \mathcal{T}_0^c}\mathbb{E} \sum_{k = 1}^{t} [(1 - \alpha) V^{\pi_0} - V^{\pi^t}] \notag\\
                  & = \max_{t \in \mathcal{T}_0^c}\mathbb{E} \sum_{k = 1}^{t} [V^{*} - V^{\pi^t} - \alpha V^{\pi_0} - (V^* - V^{\pi_0})] \notag\\
                  & =  \max_{t \in \mathcal{T}_0^c} \mathbb{E}[R^{T_0^c}_{\mathfrak{A}}(\mathcal{A}_{c})(t)] - (\alpha V^{\pi_0} + \Delta_0) t,
            \end{align}
            where $\Delta_0 = V^* - V^{\pi_0}$ is the difference between the optimal policy and the baseline policy and $\mathbb{E}[R^{T_0^c}_{\mathfrak{A}}(\mathcal{A}_{c})(t)]$ is the regret incurred by the rounds $\{i_k\}_{k \in [t]}$.
            Therefore, for any $t \in [|\mathcal{T}_0^c|]$, by Eq.~\eqref{eq:200} we have that there exists an instance $u$ (for instance in a bandit problem
            $u$ is the means of each arm) of $\mathfrak{A}$ such that $\mathbb{E}[R^{T_0^c}_{\mathfrak{A}}(\mathcal{A}_{c})(t)] \geq \xi\sqrt{t}$.  
            Let $t_0 = \frac{\xi^2}{4(\alpha V^{\pi_0} + \Delta_0)^2}$, then there exists an instance such that 
            \begin{align}
                B_{\mathcal{T}_0^c}(\mathcal{A}_{c}) \geq  \xi \sqrt{t_0} - (\alpha V^{\pi_0} + \Delta_0) t_0 \gtrsim  \frac{\xi^2}{\alpha V^{\pi_0} + \Delta_0}.
            \end{align}
            Combining the conservative condition in Equation \eqref{eq:conservative.condition}, we have
          \begin{align*}
      \mathbb{E}|\mathcal{T}_0| \geq \frac{B_{\mathcal{T}_0^c}(\mathcal{A}_{c})}{\alpha V^{\pi_0}} \gtrsim \frac{\xi^2}{\alpha V^{\pi_0} \cdot (\alpha V^{\pi_0} + \Delta_0)}.
  \end{align*}
  By the same derivation of Equation \eqref{eq:201}, we have
  \begin{align} \label{eq:204}
      \mathbb{E}[R^{T}_{\mathfrak{A}}(\mathcal{A}_{c})] \gtrsim \frac{\xi^2 \Delta_0}{\alpha V^{\pi_0} \cdot (\alpha V^{\pi_0} + \Delta_0)}.
  \end{align}
        \end{itemize}
    }

  Combining Equations \eqref{eq:200}, \eqref{eq:201}, and \eqref{eq:204}, we obtain
  \begin{align}
      \mathbb{E}[R^{T}_{\mathfrak{A}}(\mathcal{A})] \gtrsim \max \Big\{\xi\sqrt{T}, \frac{\xi^2 \Delta_0}{\alpha V^{\pi_0} \cdot (\alpha V^{\pi_0} + \Delta_0)}\Big\}.
  \end{align}
  Then we discuss the lower bound for different setups.
  \begin{itemize}
      \item For multi-armed bandits, by Lemma \ref{lemma:lb:mab}, we choose $\xi = \sqrt{K}$. Then we have
      \begin{align*}
      \mathbb{E}[R_T] \gtrsim \max \Big\{\sqrt{KT}, \frac{\xi^2 \Delta_0}{\alpha V^{\pi_0} \cdot (\alpha V^{\pi_0} + \Delta_0)}\Big\}.
      \end{align*}
      \item For linear bandits, by Lemma \ref{lemma:lb:lb}, we choose $\xi = d$. Then we have
      \begin{align*}
      \mathbb{E}[R_T] \gtrsim \max \Big\{d\sqrt{T}, \frac{d^2 \Delta_0}{\alpha V^{\pi_0} \cdot (\alpha V^{\pi_0} + \Delta_0)}\Big\}.
      \end{align*}
      \item For tabular RL, by Lemma \ref{lemma:lb:tabular}, we choose $\xi = \sqrt{SAH^3}$. Then we have
      \begin{align*}
      \mathbb{E}[R_T] \gtrsim \max \Big\{\sqrt{SAH^3T}, \frac{SAH^3 \Delta_0}{\alpha V^{\pi_0} \cdot (\alpha V^{\pi_0} + \Delta_0)}\Big\}.
      \end{align*}
      \item For low-rank MDP, by Lemma \ref{lemma:lb:linmdp}, we choose $\xi = \sqrt{d^2H^3}$. Then we have
      \begin{align*}
      \mathbb{E}[R_T] \gtrsim \max \Big\{\sqrt{d^2H^3T}, \frac{d^2H^3 \Delta_0}{\alpha V^{\pi_0} \cdot (\alpha V^{\pi_0} + \Delta_0)}\Big\}.
      \end{align*}
  \end{itemize}
  Therefore, we conclude the proof.
\end{proof}